\documentclass[11pt]{article}

\usepackage[final]{acl}

\usepackage{times}
\usepackage{latexsym}

\usepackage[T1]{fontenc}

\usepackage[utf8]{inputenc}

\usepackage{microtype}

\usepackage{inconsolata}

\usepackage{graphicx}

\usepackage{multirow}
\usepackage{enumitem}
\usepackage[dvipsnames]{xcolor}
\usepackage[table]{xcolor}
\usepackage{tabularx}
\usepackage[most]{tcolorbox}
\usepackage{bm}
\usepackage{setspace}
\usepackage{booktabs}
\usepackage{pifont}
\usepackage{caption}
\usepackage{xspace}
\usepackage{xcolor}

\definecolor{myred}{RGB}{178,34,34}
\definecolor{mygreen}{RGB}{34,178,34}
\definecolor{mygray}{RGB}{245,245,245}
\newcommand{\cmark}{{\color{mygreen} \ding{51}}}
\newcommand{\xmark}{{\color{myred} \ding{55}}}
\newcommand{\ours}{\textsc{SciImpact}\xspace}
\newcommand{\artifact}{\mathcal{A}}
\newcommand{\modelname}[1]{\texttt{#1}}

\title{\ours: A Multi-Dimensional, Multi-Field Benchmark for \\ Scientific Impact Prediction}

\author{
Hangxiao Zhu\textsuperscript{1}, 
Yuyu Zhang\textsuperscript{2}, 
Ping Nie\textsuperscript{3}, 
Yu Zhang\textsuperscript{1} \\
\textsuperscript{1}Texas A\&M University \quad 
\textsuperscript{2}Verdent AI \quad 
\textsuperscript{3}University of Waterloo \\
\texttt{\{hangxiao, yuzhang\}@tamu.edu} \\
}

\begin{document}
\maketitle
\begin{spacing}{1}
\begin{abstract}
The rapid growth of scientific literature calls for automated methods to assess and predict research impact.
Prior work has largely focused on citation-based metrics, leaving limited evaluation of models’ capability to reason about other impact dimensions.
To this end, we introduce \ours, a large-scale, multi-dimensional benchmark for scientific impact prediction spanning 19 fields.
\ours captures various forms of scientific influence, ranging from citation counts to award recognition, media attention, patent reference, and artifact adoption, by integrating heterogeneous data sources and targeted web crawling.
It comprises 215,928 contrastive paper pairs reflecting meaningful impact differences in both short- (e.g., Best Paper Award) and long-term settings (e.g., Nobel Prize).
We evaluate 11 widely used large language models (LLMs) on \ours.
Results show that off-the-shelf models show substantial variability across dimensions and fields, while multi-task supervised fine-tuning consistently enables smaller LLMs (e.g., 4B) to markedly outperform much larger models (e.g., 30B) and surpass powerful closed-source LLMs (e.g., \modelname{o4-mini}).
These results establish \ours as a challenging benchmark and demonstrate its value for multi-dimensional, multi-field scientific impact prediction.
Our project homepage is \url{https://flypig23.github.io/sciimpact-homepage/}.
\end{abstract}

\section{Introduction}
As scientific literature continues to grow exponentially \cite{dong2017century}, researchers face unprecedented challenges in identifying influential research from an ever-expanding body of work. This challenge motivates the development of techniques for predicting which studies are likely to become influential in the future \cite{xia2023review}, thereby supporting effective knowledge acquisition, scientific evaluation, and decision-making. Prior work on scientific impact prediction has largely focused on citation count prediction and its variants \cite{dong2015will,li2019neural,hirako2023realistic}. However, while citations are positively correlated with some other measures of scientific recognition \cite{jin2021scientific,zhang2025internal}, they alone are insufficient to capture the full range of factors that reflect impact \cite{radicchi2017quantifying}. In particular, the following aspects also warrant consideration.

\begin{figure}[t]
  \centering
  \includegraphics[width=\linewidth]{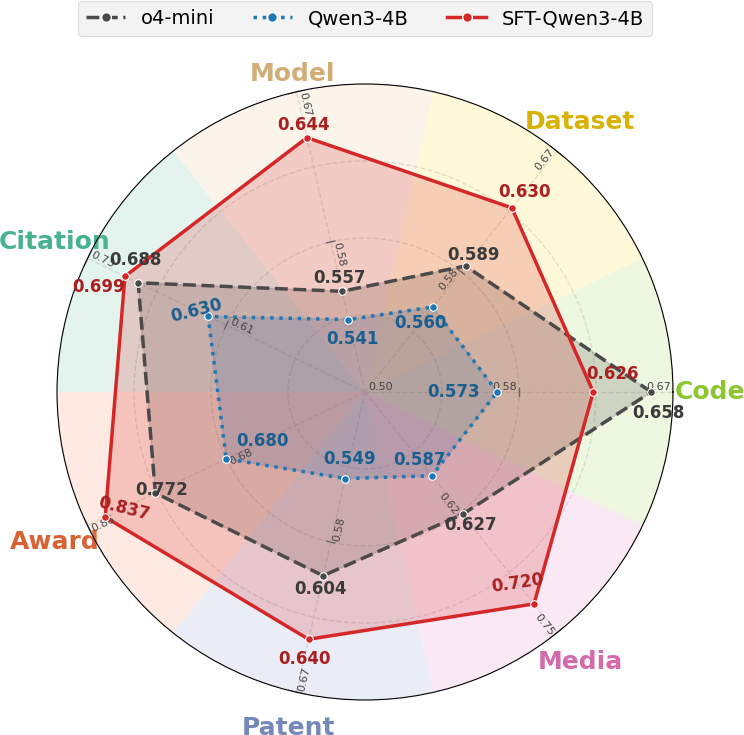}
  \caption{Performance of \modelname{o4-mini}, off-the-shelf \modelname{Qwen3-4B}, and supervised fine-tuned \modelname{Qwen3-4B} across the seven impact dimensions on \ours.
  Supervised fine-tuning (SFT) substantially enhances a 4B open-weight model's ability to predict scientific impact across all dimensions, enabling it to rival or surpass a stronger closed-source model.}
  \label{fig:radar_metric}
  \vspace{-0.5em}
\end{figure}

\begin{table*}[t]
\centering
\small
\scalebox{0.88}{
\begin{tabular}{l|ccccccc|ccc}
\toprule
& \multicolumn{7}{c|}{\textbf{Dimension Coverage}} & \multicolumn{3}{c}{\textbf{Field Coverage}} \\
\midrule
& Citation & Award & Patent & Media & Code & Dataset & Model & Comp. Sci. & Biomedicine & Other Fields \\
\midrule
\citet{li2019dataset} & \xmark & \cmark & \xmark & \xmark & \xmark & \xmark & \xmark & \xmark & \cmark & \cmark \\
\citet{li2019neural} & \cmark & \xmark & \xmark & \xmark & \xmark & \xmark & \xmark & \cmark & \xmark & \xmark \\
\citet{hirako2023realistic} & \cmark & \xmark & \xmark & \xmark & \xmark & \xmark & \xmark & \cmark & \cmark & \xmark \\
\citet{lin2023sciscinet} & \cmark & \cmark & \cmark & \cmark & \xmark & \xmark & \xmark & \cmark & \cmark & \cmark \\
\citet{yang2024navigating} & \xmark & \xmark & \xmark & \xmark & \xmark & \cmark & \xmark & \cmark & \xmark & \xmark \\
\citet{liang2024systematic} & \xmark & \xmark & \xmark & \xmark & \xmark & \xmark & \cmark & \cmark & \xmark & \xmark \\
\citet{zhang2025internal} & \cmark & \xmark & \cmark & \cmark & \cmark & \xmark & \xmark & \cmark & \xmark & \xmark \\
\midrule
\rowcolor{teal!8}
\ours (Ours) & \cmark & \cmark & \cmark & \cmark & \cmark & \cmark & \cmark & \cmark & \cmark & \cmark \\
\bottomrule
\end{tabular}
}
\caption{Comparison between \ours and existing data sources.}
\vspace{-0.5em}
\label{tab:comparison}
\end{table*}

\smallskip
\noindent \textbf{Award Recognition.} Prize-winning topics produce 47\% more star scientists and attract 37\% more new entrants \cite{jin2021scientific}. In physics, chemistry, medicine, and economics, predicting which papers may lead their authors to win a Nobel Prize is an annually high-profile task closely related to impact prediction. In computer science conferences, best paper award prediction offers another perspective on academic impact \cite{huang2023best}.

\smallskip
\noindent \textbf{Public Use.} Scientific articles are not only cited within the ``ivory tower'' of academia but are also consumed in public domains, such as technological outlets (e.g., patents) and societal channels (e.g., news and social media). Previous studies \cite{yin2022public,zhang2025internal} have shown that papers referenced in patents or media posts are 5 to 18 times as likely to become high-impact compared to a randomly selected paper.

\smallskip
\noindent \textbf{Artifact Adoption.} Scientific papers, especially in computer science, are often accompanied by artifacts such as codebases \cite{paperswithcode}, constructed datasets \cite{yang2024navigating}, and pre-trained models \cite{liang2024systematic} hosted on platforms like GitHub or Hugging Face. Intuitively, the number of times these byproducts are downloaded or starred by users on such platforms also serves as a crucial measure of a paper's impact.

\smallskip
To bridge the gap between prior work predominantly targeting citation count prediction and the multi-faceted impact criteria outlined above, in this paper, we propose \ours, a comprehensive, multi-dimensional, and multi-field benchmark for scientific impact evaluation.
As shown in Table \ref{tab:comparison}, \ours covers 7 distinct impact dimensions (Citation, Award, Patent, Media, Code, Dataset, and Model), strictly more than any single existing data source to the best of our knowledge. Moreover, \ours goes beyond computer science and biomedicine papers emphasized in previous scientific literature understanding studies, encompassing papers from natural sciences, engineering, social sciences, and humanities (corresponding to all 19 fields in the Microsoft Academic Graph \cite{shen2018web}). This results in 215,928 contrastive paper pairs, enabling pairwise impact prediction in which models determine which paper in each pair has greater impact in a given dimension.\footnote{In this paper, we cast the task of ``prediction'' as binary classification, following prior comparative formulations in \citet{sayyadi2009futurerank} and \citet{dong2015will}.}
It is worth noting that, in constructing this benchmark, we not only curate data from fragmented and heterogeneous existing resources but also crawl missing data for specific dimensions and fields from the web (e.g., MDPI Best Paper Awards and GitHub star counts).

Based on \ours, we conduct a comprehensive evaluation of 11 prominent large language models (LLMs) for scientific impact prediction, including 3 closed-source models and 8 open-source models. In addition, we aggregate training data across all impact dimensions and perform multi-task instruction tuning to train unified scientific impact prediction models using \modelname{Qwen3-4B} \cite{yang2025qwen3} and \modelname{LLaMA-3.2-3B} \cite{grattafiori2024llama} as backbones. Figure~\ref{fig:radar_metric} compares a representative closed-source model (\modelname{o4-mini}), an off-the-shelf open-weight model (\modelname{Qwen3-4B}), and its supervised fine-tuned counterpart (\modelname{SFT-Qwen3-4B}) across the seven impact dimensions, illustrating the benefits of fine-tuning on \ours. (We provide the corresponding comparison across fields for the same three models in Appendix~\ref{app:radar_field}.) Overall, our results show that the fine-tuned 4B model delivers the strongest average performance and is competitive with leading closed-source baselines, underscoring the value of \ours for scientific impact prediction as a multi-dimensional task.

The contributions of our work are as follows:
\begin{itemize}[leftmargin=*]
\item We broaden the scope of scientific impact prediction by framing impact as a multi-dimensional concept that goes beyond citation counts, incorporating diverse forms of award recognition, public use, and artifact adoption.

\item To support this perspective, we introduce \ours, a comprehensive, multi-dimensional, multi-field benchmark for scientific impact evaluation, built via curated integration of fragmented, heterogeneous resources and targeted web crawling to fill missing dimensions and fields.

\item We conduct large-scale experiments on \ours to evaluate various prominent LLMs and train unified scientific impact prediction models via multi-task instruction tuning. Results show that this fine-tuning enables relatively small LLMs to achieve superior performance compared to much larger or stronger models.

\end{itemize}

\section{Related Work}

\subsection{Evolution of Scientific Impact Prediction}

Quantitative studies of scientific literature primarily use citation counts as a proxy for impact \cite{wang2013quantifying,sinatra2016quantifying}. Early work predicts future citations from features available at or shortly after publication, such as author history and bibliometric cues \cite{castillo2007estimating,fu2008models,ibanez2009predicting}. Later studies reveal that heterogeneous citation trajectories, motivating dynamic models that account for temporal effects including aging and cumulative advantage \cite{chakraborty2014towards,xiao2016modeling}. With the advent of deep learning, sequence-based approaches further improve citation forecasting by modeling temporal dependencies \cite{yuan2018modeling}. Auxiliary signals, such as peer review text, also proved to enhance prediction \cite{li2019neural}. More recently, LLMs have been applied to citation prediction tasks \cite{zhao2025words,lu2025newborn,zhang2024comprehensive} due to their strong understanding and reasoning capabilities.


Beyond citations, impact includes artifact adoption and external attention. Work analyzes popularity signals in open-source ecosystems, such as GitHub stars and their relationship to downstream usage \cite{ren2020starin,koch2024fault}, as well as dataset and model reuse on platforms like Hugging Face \cite{koch2021reduced,yang2024navigating,liang2024systematic}. External influence is also quantified using patents, media, and policy documents alongside scholarly citations \cite{yin2022public,zhang2025internal}.


Overall, existing studies typically focus on a single proxy, platform, or prediction horizon, and lack a unified benchmark for systematic comparison across fields and impact dimensions. These limitations motivate \ours, which provides standardized evaluation over diverse disciplines and heterogeneous indicators of scientific influence.

\subsection{Datasets for Science Literature Analysis}


Advances in science literature analysis are enabled by large-scale scholarly datasets. Early work widely relies on the Microsoft Academic Graph (MAG; \citealp{shen2018web}), which also supports curated resources such as Nobel-laureate publication datasets \cite{li2019dataset}. Following MAG's discontinuation, OpenAlex \cite{priem2022openalex} emerges as a fully open alternative with broad metadata and citation coverage. Complementary resources improve cross-disciplinary analysis \cite{gao2025science}, including MAPLE for field-aware topic tagging \cite{zhang2023effect} and SciSciNet as an integrated data lake linking publications to external signals \cite{lin2023sciscinet}.
Recent datasets are often released with task-specific benchmarks, such as impact forecasting on evolving scholarly graphs \cite{gu2024impact4cast}, interdisciplinary link prediction \cite{rezaee2025fos}, and text impact prediction for newborn papers using LLMs \cite{zhao2025words}. While these resources advance metascience studies, they typically center on a single task or dataset family. \ours complements them by providing a unified benchmark for scientific impact prediction across multiple fields and dimensions.

\section{\ours Benchmark}

\begin{figure*}[t]
\centering
\includegraphics[width=\textwidth]{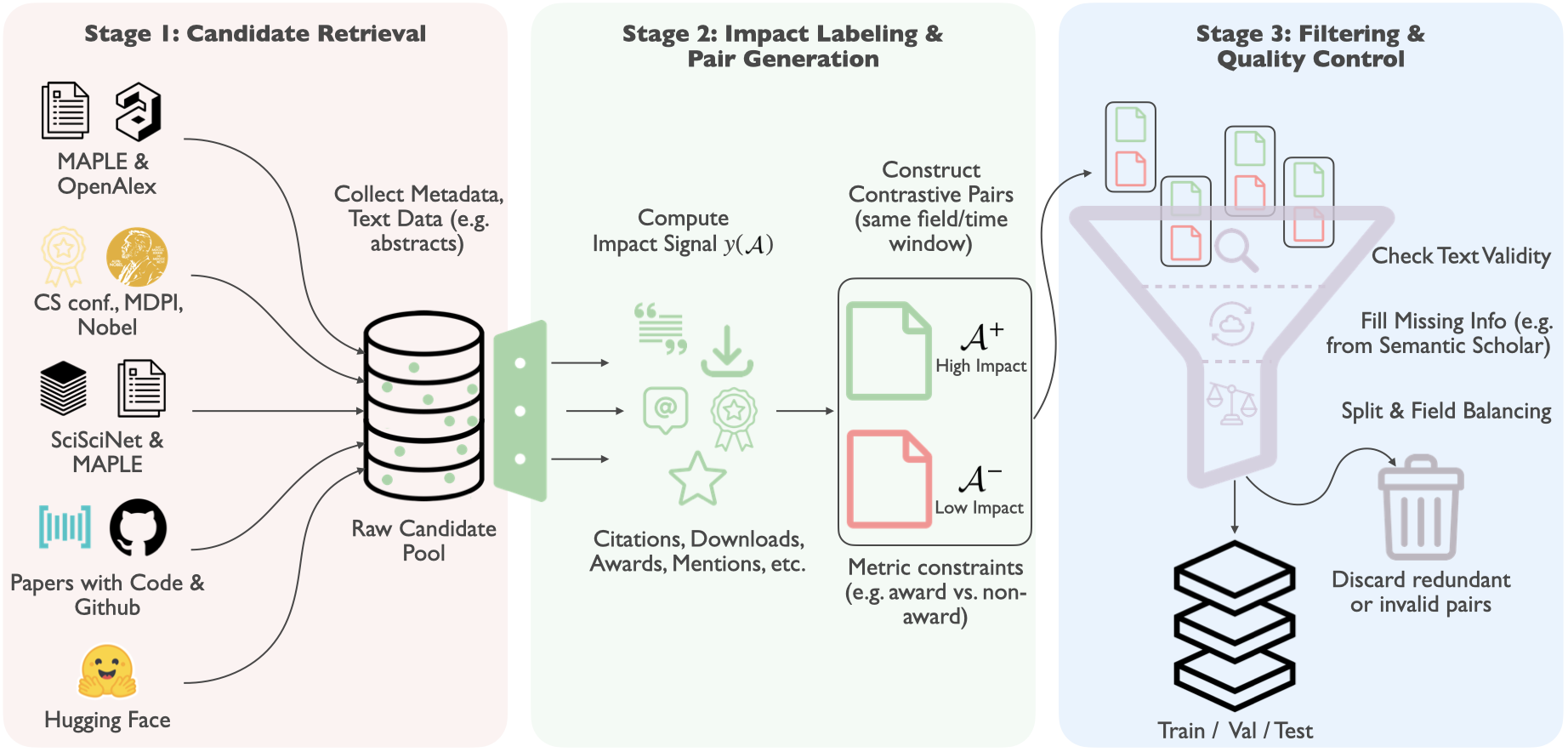}
\caption{Overview of the \ours benchmark curation pipeline, including candidate retrieval, impact labeling and pair generation, and filtering and quality control.}
\vspace{-0.5em}
\label{fig:pipeline}
\end{figure*}

We now describe the construction of our \ours benchmark.
Each instance in \ours is a \emph{contrastive} pair of artifacts $(\artifact^{+}, \artifact^{-})$, where $\artifact^{+}$ exhibits a higher impact signal than $\artifact^{-}$ within a certain dimension.
Here, ``artifacts'' may refer to research papers or their associated model cards, dataset cards, or repository README files.
\ours covers 19 fields, spanning art, biology, business, chemistry, computer science, economics, engineering, environmental science, geography, geology, history, materials science, mathematics, medicine, philosophy, physics, political science, psychology, and sociology.
We construct the benchmark by integrating online resources with existing datasets through a three-stage pipeline: (1) candidate retrieval, (2) impact labeling and pair generation, and (3) filtering and quality control.
Figure~\ref{fig:pipeline} summarizes the pipeline.

For the impact metric $y(\artifact)$, we consider seven dimensions that capture both academic and broader forms of influence: (1) citations; (2) award recognition (including Best Paper Awards from major computer science conferences, the Nobel Prize for physics/chemistry/medicine, and MDPI Best Paper Awards for other fields); (3) patent references; (4) media attention (combining news coverage and social media mentions); (5) GitHub stars; (6) Hugging Face dataset downloads; and (7) Hugging Face model downloads. Table~\ref{tab:impact-metrics} summarizes the thresholds applied to construct contrastive pairs for each impact dimension.

\begin{table}[t]
\centering
\small
\setlength{\tabcolsep}{6pt}
\renewcommand{\arraystretch}{1.5}
\scalebox{0.95}{
\begin{tabular}{ll}
\toprule
\textbf{Dimension} & \textbf{Pair Construction Rule} \\
\midrule
Citation &
$y(\artifact^{+}) \ge 10,\; y(\artifact^{-}) \ge 10,\; \frac{y(\artifact^{+})}{y(\artifact^{-})} \ge 2$ \\
Award &
$y(\artifact^{+})=\texttt{True},\; y(\artifact^{-})=\texttt{False}$ \\
Patent &
$y(\artifact^{+}) \ge 5,\; y(\artifact^{-}) \ge 5,\; \frac{y(\artifact^{+})}{y(\artifact^{-})} \ge 2$ \\
Media &
$y(\artifact^{+}) \ge 5,\; y(\artifact^{-}) \ge 5,\; \frac{y(\artifact^{+})}{y(\artifact^{-})} \ge 2$ \\
Code &
$y(\artifact^{+}) \ge 10,\; y(\artifact^{-}) \ge 10,\; \frac{y(\artifact^{+})}{y(\artifact^{-})} \ge 2$ \\
Dataset &
$y(\artifact^{+}) \ge 10,\; y(\artifact^{-}) \ge 10,\; \frac{y(\artifact^{+})}{y(\artifact^{-})} \ge 2$ \\
Model &
$y(\artifact^{+}) \ge 10,\; y(\artifact^{-}) \ge 10,\; \frac{y(\artifact^{+})}{y(\artifact^{-})} \ge 2$ \\
\bottomrule
\end{tabular}
}
\caption{Impact dimensions and thresholding rules used to construct contrastive pairs in \ours.
\textbf{Note:} For award recognition, $y(\artifact)$ is a boolean indicator reflecting whether the artifact receives the corresponding award. For all other dimensions, $y(\artifact)$ is a nonnegative count (e.g., citation count).}
\vspace{-0.5em}
\label{tab:impact-metrics}
\end{table}

\subsection{Stage 1: Candidate Retrieval}
\label{sec:candidate-retrieval}

\smallskip
\noindent \textbf{Citation.}
We first retrieve candidate papers from MAPLE \cite{zhang2023effect}, which collects research articles published in the top-100 venues of each of the 19 fields.
Publication years are restricted to 2001-2020 to allow sufficient time for citations to accumulate.
We then match MAPLE entries with OpenAlex \cite{priem2022openalex} to obtain the title, abstract, year, and citation count up to mid-2025, which serves as $y(\artifact)$.
To ensure a fair comparison, $\artifact^{+}$ and $\artifact^{-}$ in a contrastive pair are required to be published in the same year.
Note that the Citation dimension encompass scientific impact prediction over different time horizons: pairs published in 2001 correspond to longer-term impact prediction, while pairs from 2020 represent a shorter-term prediction horizon.

\smallskip
\noindent \textbf{Award.}
We crawl award data from three sources depending on the field: (1) Best Paper Awards from major computer science conferences \cite{huang2023best}, (2) Nobel Prize-winning papers for physics, chemistry, and medicine \cite{li2019dataset}, and (3) MDPI Best Paper Awards for the remaining fields \cite{mdpi}.
We link each award-winning paper to OpenAlex via DOI matching and collect the required bibliographic metadata.
We set $y(\artifact)=\texttt{True}$ for award-winning papers and sample corresponding non-award-winning papers with $y(\artifact)=\texttt{False}$.
To be specific, for Best Paper Awards, the non-award-winning paper $\artifact^{-}$ is required to be published in the same venue as the award-winning paper $\artifact^{+}$.
For the Nobel Prize, $\artifact^{-}$ is required to be authored by the same scientist as $\artifact^{+}$, ensuring comparability within an author's body of work.
Note that the Award dimension also spans different time horizons: Best Paper Award prediction corresponds to a shorter-time horizon, as such awards are typically announced within a few months after paper acceptance. In contrast, the Nobel Prize reflects longer-term impact, given the substantial time lag between publication and the conferral of the award \cite{mitsis2022nobel}.

\smallskip
\noindent \textbf{Patent and Media.}
We retrieve the records of papers referenced by patents and news/social media posts from SciSciNet \cite{lin2023sciscinet}.
For other public-use dimensions, such as policy documents, the corresponding resources \cite{szomszor2022overton} require restricted access and are not publicly available.
Therefore, we do not include them in \ours.
We link SciSciNet papers to MAPLE using the MAG identifier to determine the field (among the 19) to which each paper belongs.
$y(\artifact)$ is defined as the number of patent references and media mentions, respectively, recorded by SciSciNet.

\smallskip
\noindent \textbf{Code.} We retrieve paper-associated GitHub repositories from Papers with Code \cite{paperswithcode}, retaining those with at least 10 stars and discarding repositories with missing or extremely short README files. We collect the star count of each retrieved repository via the GitHub REST API \cite{github}, which defines $y(\artifact)$. The repository README serves as the primary textual input for the artifact $\artifact$.

\smallskip
\noindent \textbf{Dataset and Model.} To capture adoption in the machine learning ecosystem, we retrieve Hugging Face dataset and model cards from \citet{yang2024navigating} and \citet{liang2024systematic}, respectively. For each artifact, we collect the card text and platform-provided statistics, defining $y(\artifact)$ as the corresponding download count.

\subsection{Stage 2: Impact Labeling and Pair Generation}
\label{sec:labeling-pair-generation}

In Stage 2, after computing the impact metric $y(\artifact)$ for each candidate retrieved in Stage 1, we construct contrastive pairs $(\artifact^{+}, \artifact^{-})$ within each dimension and field. Following the dimension-specific constraints in Table~\ref{tab:impact-metrics}, each pair is formed by selecting two artifacts from the same field (and matching publication year, venue, or author when applicable, as described in Section~\ref{sec:candidate-retrieval}) such that $\artifact^{+}$ exhibits higher impact than $\artifact^{-}$. For count-based dimensions, both artifacts must exceed a minimum activity threshold and satisfy a minimum relative gap (e.g., $y(\artifact^{+})/y(\artifact^{-}) \ge 2$) to ensure meaningful contrast.

\subsection{Stage 3: Filtering and Quality Control}
\label{sec:filtering-qc}
In Stage 3, we filter and sample the constructed pairs to improve text completeness, reduce noise, and balance coverage across fields. We prioritize pairs with complete textual inputs required for modeling (e.g., title and abstract for papers) and discard candidates with missing or clearly invalid text. To balance fields, we target 4,000/3,000/3,000 train/validation/test pairs for computer science, physics, chemistry, and medicine, and 400/300/300 for each remaining field. If a field lacks enough qualified pairs, we retain all available ones. For pairs with missing text, we attempt recovery by re-fetching from online resources (e.g., Semantic Scholar; \citealp{ammar2018construction}) using identifiers or title-based matching, keeping only reliably recovered text. Finally, we remove duplicate pairs induced by cross-source linking.

\subsection{Dataset Statistics}

\begin{table}[t]
\centering
\small
\setlength{\tabcolsep}{3pt}
\renewcommand{\arraystretch}{1.15}
\scalebox{0.95}{
\begin{tabular}{lccc}
\toprule
\textbf{Dimension} & \textbf{\# of Pairs} & \textbf{Mean Text Len.} & \textbf{Mean Pair Len.} \\
\midrule
Citation & 43,309 & 156.9 & 313.8 \\
Award   & 42,033 & 114.8 & 229.6 \\
Patent  & 45,745 & 160.2 & 320.5 \\
Media   & 52,739 & 166.8 & 333.6 \\
Code    & 9,193  & 448.9 & 897.8 \\
Dataset & 10,517 & 344.8 & 689.5 \\
Model   & 12,463 & 257.6 & 515.2 \\
\bottomrule
\end{tabular}
}
\caption{Dataset statistics by dimension. Mean lengths are measured in word count per artifact input and per contrastive pair $(\artifact^{+}, \artifact^{-})$, respectively.}
\label{tab:dataset-stats}
\end{table}

\begin{figure}[t]
\centering
\includegraphics[width=0.95\columnwidth]{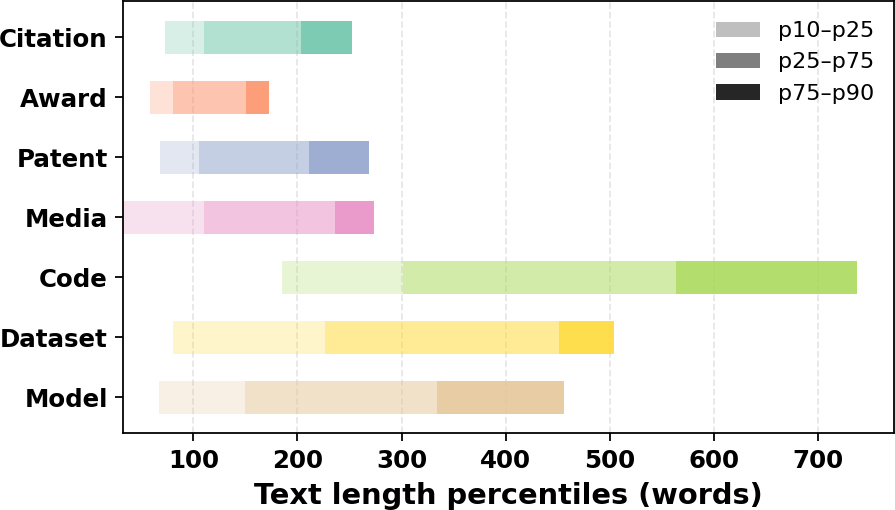}
\caption{Text length distribution by dimension. Each horizontal bar represents percentile ranges of artifact input length (word count): p10–p25 (light), p25–p75 (medium), and p75–p90 (dark).}
\vspace{-0.5em}
\label{fig:textlen-percentiles}
\end{figure}

Table~\ref{tab:dataset-stats} summarizes the number of contrastive pairs and the average input length (word count) for each dimension. One can observe that the mean artifact text length varies across dimensions: tasks using paper abstracts (i.e., Citation, Award, Patent, and Media) exhibit similar lengths, as abstracts are typically concise (often under 300 words). By contrast, tasks using repository or card text (i.e., Code, Dataset, and Model) have longer and more variable inputs due to the richer, heterogeneous content in README files and Hugging Face cards. Figure~\ref{fig:textlen-percentiles} illustrates these patterns with percentile bands of text length for each impact dimension.

\section{Experiments}

\begin{table*}[t]
\centering
\small
\renewcommand{\arraystretch}{1.2}
\scalebox{0.88}{
\begin{tabular}{lcccccccl}
\toprule
 & \textbf{Citation} & \textbf{Award} & \textbf{Patent} & \textbf{Media} & \textbf{Code} & \textbf{Dataset} & \textbf{Model} & \cellcolor{black!10} \textbf{Average} \\
\midrule
\rowcolor[HTML]{E6F3FF}
\multicolumn{9}{c}{\textbf{Closed-Source Models}} \\
\modelname{GPT-4.1-mini} \cite{achiam2023gpt} & 0.664 & 0.745 & 0.592 & 0.608 & 0.603 & 0.596 & 0.572 & \cellcolor{black!10} 0.626$^{**}$ \\
\modelname{o4-mini} \cite{openai2025o3o4mini} & 0.688 & 0.772 & 0.604 & 0.627 & \textbf{0.658} & 0.589 & 0.557 & \cellcolor{black!10} 0.642$^{*}$ \\
\modelname{Claude-haiku-4.5} \cite{anthropic2025claudehaiku45} & 0.662 & 0.780 & 0.596 & 0.632 & 0.626 & 0.602 & 0.542 & \cellcolor{black!10} 0.634$^{**}$ \\
\midrule
\rowcolor[HTML]{E6FFEA}
\multicolumn{9}{c}{\textbf{Open-Source Models}} \\
\modelname{LLaMA-3.2-3B} \cite{grattafiori2024llama} & 0.534 & 0.539 & 0.534 & 0.517 & 0.513 & 0.526 & 0.548 & \cellcolor{black!10} 0.530$^{**}$ \\
\modelname{LLaMA-3-8B} \cite{grattafiori2024llama} & 0.552 & 0.625 & 0.534 & 0.594 & 0.547 & 0.549 & 0.534 & \cellcolor{black!10} 0.562$^{***}$ \\
\modelname{LLaMA-3.1-8B} \cite{grattafiori2024llama} & 0.579 & 0.652 & 0.534 & 0.589 & 0.525 & 0.534 & 0.535 & \cellcolor{black!10} 0.564$^{***}$ \\
\modelname{Qwen3-4B} \cite{yang2025qwen3} & 0.630 & 0.680 & 0.549 & 0.587 & 0.573 & 0.560 & 0.541 & \cellcolor{black!10} 0.589$^{***}$ \\
\modelname{Qwen2.5-7B} \cite{yang2025qwen2} & 0.601 & 0.646 & 0.557 & 0.604 & 0.563 & 0.592 & 0.560 & \cellcolor{black!10} 0.589$^{**}$ \\
\modelname{Qwen2.5-14B} \cite{yang2025qwen2} & 0.565 & 0.672 & 0.586 & 0.620 & 0.577 & 0.561 & 0.562 & \cellcolor{black!10} 0.592$^{**}$ \\
\modelname{Ministral-3-3B} \cite{mistral2025ministral3} & 0.559 & 0.642 & 0.536 & 0.607 & 0.542 & 0.503 & 0.519 & \cellcolor{black!10} 0.558$^{***}$ \\
\modelname{Nemotron-3-Nano-30B} \cite{blakeman2025nemotron}& 0.537 & 0.618 & 0.500 & 0.504 & 0.528 & 0.565 & 0.549 & \cellcolor{black!10} 0.543$^{***}$ \\
\midrule
\rowcolor[HTML]{FFF5E6}
\multicolumn{9}{c}{\textbf{Fine-Tuned Models}} \\
\modelname{SFT-LLaMA-3.2-3B} & 0.653 & 0.806 & 0.629 & 0.697 & 0.618 & 0.618 & 0.625 & \cellcolor{black!10} 0.664$^{**}$ \\
\modelname{SFT-Qwen3-4B} & \textbf{0.699} & \textbf{0.837} & \textbf{0.640} & \textbf{0.720} & 0.626 & \textbf{0.630} & \textbf{0.644} & \cellcolor{black!10} \textbf{0.685} \\
\bottomrule
\end{tabular}
}
\caption{Pairwise prediction accuracy of different models across 7 impact dimensions. The \textbf{Average} column reports the average performance across all dimensions. Bold values denote the best score within each dimension. Asterisks indicate statistical significance compared to \modelname{SFT-Qwen3-4B} ($^{*}: p < 0.05$, $^{**}: p < 0.01$, $^{***}: p < 0.001$).}
\vspace{-0.5em}
\label{tab:metric_results}
\end{table*}

\begin{table*}[t]
\centering
\small
\renewcommand{\arraystretch}{1.2}
\scalebox{0.88}{
\begin{tabular}{lcccccl}
\toprule
 & \textbf{Comp. Sci.} & \textbf{Physics} & \textbf{Chemistry} & \textbf{Medicine} & \textbf{Other Fields} & \cellcolor{black!10} \textbf{Average} \\
\midrule
\rowcolor[HTML]{E6F3FF}
\multicolumn{7}{c}{\textbf{Closed-Source Models}} \\
\modelname{GPT-4.1-mini} \cite{achiam2023gpt} & 0.625 & 0.706 & 0.685 & 0.673 & 0.586 & \cellcolor{black!10} 0.655$^{*}$ \\
\modelname{o4-mini} \cite{openai2025o3o4mini} & 0.639 & \textbf{0.730} & 0.690 & 0.710 & 0.617 & \cellcolor{black!10} 0.677 \\
\modelname{Claude-haiku-4.5} \cite{anthropic2025claudehaiku45} & 0.631 & 0.680 & 0.694 & 0.730 & 0.615 & \cellcolor{black!10} 0.670$^{*}$ \\
\midrule
\rowcolor[HTML]{E6FFEA}
\multicolumn{7}{c}{\textbf{Open-Source Models}} \\
\modelname{LLaMA-3.2-3B} \cite{grattafiori2024llama} & 0.533 & 0.541 & 0.528 & 0.525 & 0.531 & \cellcolor{black!10} 0.532$^{***}$ \\
\modelname{LLaMA-3-8B} \cite{grattafiori2024llama} & 0.561 & 0.616 & 0.585 & 0.556 & 0.551 & \cellcolor{black!10} 0.574$^{**}$ \\
\modelname{LLaMA-3.1-8B} \cite{grattafiori2024llama} & 0.560 & 0.644 & 0.607 & 0.552 & 0.559 & \cellcolor{black!10} 0.584$^{**}$ \\
\modelname{Qwen3-4B} \cite{yang2025qwen3} & 0.590 & 0.653 & 0.633 & 0.607 & 0.577 & \cellcolor{black!10} 0.612$^{**}$ \\
\modelname{Qwen2.5-7B} \cite{yang2025qwen2} & 0.587 & 0.661 & 0.617 & 0.579 & 0.565 & \cellcolor{black!10} 0.602$^{**}$ \\
\modelname{Qwen2.5-14B} \cite{yang2025qwen2} & 0.597 & 0.684 & 0.594 & 0.597 & 0.585 & \cellcolor{black!10} 0.612$^{*}$ \\
\modelname{Ministral-3-3B} \cite{mistral2025ministral3} & 0.547 & 0.619 & 0.616 & 0.577 & 0.556 & \cellcolor{black!10} 0.583$^{***}$ \\
\modelname{Nemotron-3-Nano-30B} \cite{blakeman2025nemotron} & 0.544 & 0.525 & 0.578 & 0.533 & 0.510 & \cellcolor{black!10} 0.538$^{***}$ \\
\midrule
\rowcolor[HTML]{FFF5E6}
\multicolumn{7}{c}{\textbf{Fine-Tuned Models}} \\
\modelname{SFT-LLaMA-3.2-3B} & 0.652 & 0.700 & 0.718 & 0.722 & 0.681 & \cellcolor{black!10} 0.695$^{*}$ \\
\modelname{SFT-Qwen3-4B} & \textbf{0.669} & 0.717 & \textbf{0.768} & \textbf{0.743} & \textbf{0.704} & \cellcolor{black!10} \textbf{0.720} \\
\bottomrule
\end{tabular}
}
\caption{Pairwise prediction accuracy of different models across scientific fields. The \textbf{Average} column reports the average performance across all fields. Bold values denote the best score within each field. Asterisks indicate statistical significance compared to \modelname{SFT-Qwen3-4B} ($^{*}: p < 0.05$, $^{**}: p < 0.01$, $^{***}: p < 0.001$).}
\vspace{-0.5em}
\label{tab:field_results}
\end{table*}

After constructing \ours, we comprehensively evaluate a diverse set of models on it, including:

\smallskip
\noindent \textbf{3 Closed-Source LLMs}: \modelname{GPT-4.1-mini} \cite{achiam2023gpt}, \modelname{o4-mini} \cite{openai2025o3o4mini}, and \modelname{Claude-haiku-4.5} \cite{anthropic2025claudehaiku45}

\smallskip
\noindent \textbf{8 Open-Weight LLMs}: \modelname{Qwen3-4B} \cite{yang2025qwen3}, \modelname{Qwen2.5-7B} \cite{yang2025qwen2}, \modelname{Qwen2.5- 14B} \cite{yang2025qwen2}, \modelname{LLaMA-3.2-3B} \cite{grattafiori2024llama}, \modelname{LLaMA-3-8B} \cite{grattafiori2024llama}, \modelname{LLaMA-3.1-8B} \cite{grattafiori2024llama}, \modelname{Ministral-3-3B} \cite{mistral2025ministral3}, and \modelname{Nemotron-3-Nano-30B} \cite{blakeman2025nemotron}

\smallskip
\noindent \textbf{2 Supervised Fine-tuned (SFT) Variants}: \modelname{SFT- Qwen3-4B} and \modelname{SFT-LLaMA-3.2-3B}

\subsection{Task Setup and Evaluation Protocol}
\label{sec:eval-protocol}
We use a standardized instruction-following prompt format that (1) specifies the target impact dimension and (2) constrains the output to a strict, easily parsable form. The textual input varies by dimension: for Citation, Award, Patent, and Media, we use the paper title and abstract; for Code, Dataset, and Model, we use the corresponding repository README, Hugging Face dataset card, or Hugging Face model card, respectively. Across all dimensions, inputs are truncated to a maximum of 1,000 words when necessary to ensure consistent prompt length across instances. The prompt for predicting the Best Paper Award at computer science conferences is shown below, and full prompts for all dimensions are provided in Appendix~\ref{app:prompt}.

Given two scientific artifacts $(\artifact^{+}, \artifact^{-})$ from the same field, a model is asked to predict which artifact will achieve higher future impact in a certain dimension.
(Note that in all test sets, there is a 50\% probability that option A in the prompt has higher impact than option B, and a 50\% probability of the reverse.)
We parse model outputs by exact string matching to the two allowed responses. We report \emph{pairwise accuracy}, defined as the percentage of instances in which the model correctly identifies $\artifact^{+}$ over $\artifact^{-}$.

\smallskip
\smallskip
\noindent
\colorbox{mygray}{%
\parbox{0.97\columnwidth}{%
\small
\textbf{System:} You are an impartial judge deciding which of two papers won the Best Paper Award. Your reply must be exactly one sentence and must be one of these two options: \\
\hspace*{1em}-- Paper A won the Best Paper Award \\
\hspace*{1em}-- Paper B won the Best Paper Award \\
You are not allowed to output anything else---no explanations, no extra words. \\[0.5em]
\textbf{User:} Paper A: \texttt{<artifact text for A>} \\[0.25em]
Paper B: \texttt{<artifact text for B>} \\[0.5em]
Based on the information above, which paper won the Best Paper Award? \\
Reply with exactly one sentence following the system instruction.
}%
}

\subsection{Supervised Fine-tuning Setup}
To investigate whether task-specific training improves models' performance in scientific impact prediction, we perform SFT on two representative open-weight LLMs: \modelname{LLaMA-3.2-3B} \cite{grattafiori2024llama} and \modelname{Qwen3-4B} \cite{yang2025qwen3}. Both models are fine-tuned on the training split aggregated across all impact dimensions and fields, and hyperparameters are selected based on performance on the corresponding aggregated validation split. Full-parameter fine-tuning is conducted using LLaMA-Factory\footnote{\url{https://github.com/hiyouga/LLaMA-Factory}}. All SFT experiments are performed on four NVIDIA H20 GPUs. Complete training commands and additional implementation details are provided in Appendix~\ref{app:sft-details}.

\subsection{Main Results}
Tables \ref{tab:metric_results} and \ref{tab:field_results} present the detailed performance of all models.
We also computed the averages across all dimensions and fields, and performed pairwise t-tests between each model and \modelname{SFT-Qwen3-4B}. Statistical significance is indicated in both tables.
We highlight three key observations.

\smallskip
\noindent \textbf{Effectiveness of Training on \ours.} Fine-tuned models substantially outperform their corresponding base models, showing that \ours provides a strong supervision signal for learning impact-relevant cues from artifact text. \modelname{SFT-Qwen3-4B} achieves the best performance on nearly all impact dimensions and fields; the only exceptions are Code and Physics, where \modelname{o4-mini} is the strongest. \modelname{SFT-LLaMA-3.2-3B} likewise surpasses all untuned open-weight models and several closed-source systems. Overall, relatively small open models fine-tuned on \ours can compete effectively with much larger open- and closed-source baselines. Appendix~\ref{app:encoder-baselines} reports a similar pattern for an encoder-based \modelname{SFT-SciBERT} baseline \cite{beltagy2019scibert}. Appendix~\ref{app:leakage-audit} further reports a de-leakage audit across the Code, Dataset, and Model dimensions, showing that removing explicit popularity cues from README/card text leaves performance largely unchanged.

Note that the consistent and substantial gains from SFT suggest that information memorized by LLMs during pre-training is not the dominant factor driving performance: if \ours were largely solvable by pre-training leakage or memorization, SFT would provide very limited additional benefit, whereas we observe strong and systematic improvements of SFT across models, dimensions, and fields.

\smallskip
\noindent \textbf{Complementarity Across Impact Dimensions.} To test whether different dimensions provide distinct supervision rather than merely resampling the same signal, we conduct a \emph{single-dimension SFT} ablation: for each dimension, we fine-tune a separate checkpoint using only the training data from that dimension, and then compute the accuracy of each checkpoint on its corresponding dimension. Table~\ref{tab:single_dimension_sft} shows the average accuracy across all dimensions, where single-dimension SFT improves over the untuned baseline for both Qwen and LLaMA models, but still lags behind \emph{multi-dimension SFT}, indicating complementary supervision across dimensions.

\begin{table}[t]
\centering
\small
\setlength{\tabcolsep}{6pt}
\renewcommand{\arraystretch}{1.15}
\scalebox{0.88}{
\begin{tabular}{lcc}
\toprule
 & \modelname{Qwen3-4B} & \modelname{LLaMA-3.2-3B} \\
\midrule
Untuned LLM & 0.589 & 0.530 \\
Single-Dimension SFT & 0.632 & 0.542 \\
Multi-Dimension SFT & \textbf{0.685} & \textbf{0.664} \\
\bottomrule
\end{tabular}
}
\caption{Average accuracy across all dimensions for the untuned LLM, single-dimension SFT, and multi-dimension SFT.}
\label{tab:single_dimension_sft}
\end{table}

\begin{table}[t]
\centering
\small
\setlength{\tabcolsep}{6pt}
\renewcommand{\arraystretch}{1.15}
\scalebox{0.88}{
\begin{tabular}{lcc}
\toprule
 & \textbf{Best Paper Award} & \textbf{Nobel Prize} \\
\midrule
Base Models & 0.606 & 0.737 \\
SFT Models  & 0.748 & 0.898 \\
\bottomrule
\end{tabular}
}
\caption{Average accuracy across Award subtypes. Best Paper Award aggregates the Award dimension from Computer Science and Other fields, while Nobel Prize spans Physics, Chemistry, and Medicine.
}
\vspace{-0.5em}
\label{tab:award-subtypes}
\end{table}

\smallskip
\noindent \textbf{Importance of Dimension- and Field-Specific Evaluation.} Statistical analysis using a two-factor ANOVA without replication on Tables~\ref{tab:metric_results} and~\ref{tab:field_results} reveals significant performance variation across both impact dimensions and scientific fields.

Among all dimensions, Award yields the highest prediction accuracy and is significantly easier than every other task ($p < 0.001$). One plausible explanation is that award outcomes are typically determined by a relatively small and well-defined committee rather than by diffuse, society-wide recognition accumulated over time. Compared with broader impact signals such as citations, media attention, code adoption, or dataset reuse, these committee-driven decisions may reflect a narrower set of evaluative preferences and therefore be easier for LLMs to approximate from textual cues alone. In this sense, the Award dimension may be easier not because it captures a fundamentally simpler notion of impact, but because it more often corresponds to the judgments of a bounded group of experts whose decision patterns may be more internally consistent and more learnable. Within the Award dimension, Nobel Prize related comparisons appear to be even easier than other award settings, as shown in Table~\ref{tab:award-subtypes}. One plausible explanation is that Nobel associated work in the natural sciences is more likely to contain explicit textual markers of discovery, such as named compounds or experimental protocols. In contrast, award decisions in computer science may depend more heavily on rapidly evolving trends and community dynamics that are difficult to infer from text alone. In addition, Nobel Prize related papers are often widely discussed and extensively cited, making them more likely to appear in LLM pre-training corpora. This broader exposure may confer partial prior knowledge of canonical Nobel winning contributions and further reduce the difficulty of identifying higher impact artifacts in paired comparisons.

Motivated by this field effect within the Award dimension, we further analyze field-wise differences across all impact dimensions and find that Computer Science and the aggregated Other Fields are significantly more challenging than Chemistry, Medicine, and Physics, with all $2\times3$ pairwise comparisons yielding $p < 0.01$.

These systematic variations in prediction difficulty across impact dimensions and scientific fields directly support our motivation for constructing a multi-dimensional, multi-field benchmark for scientific impact prediction.

\begin{figure}[t]
  \centering
  \includegraphics[width=\linewidth]{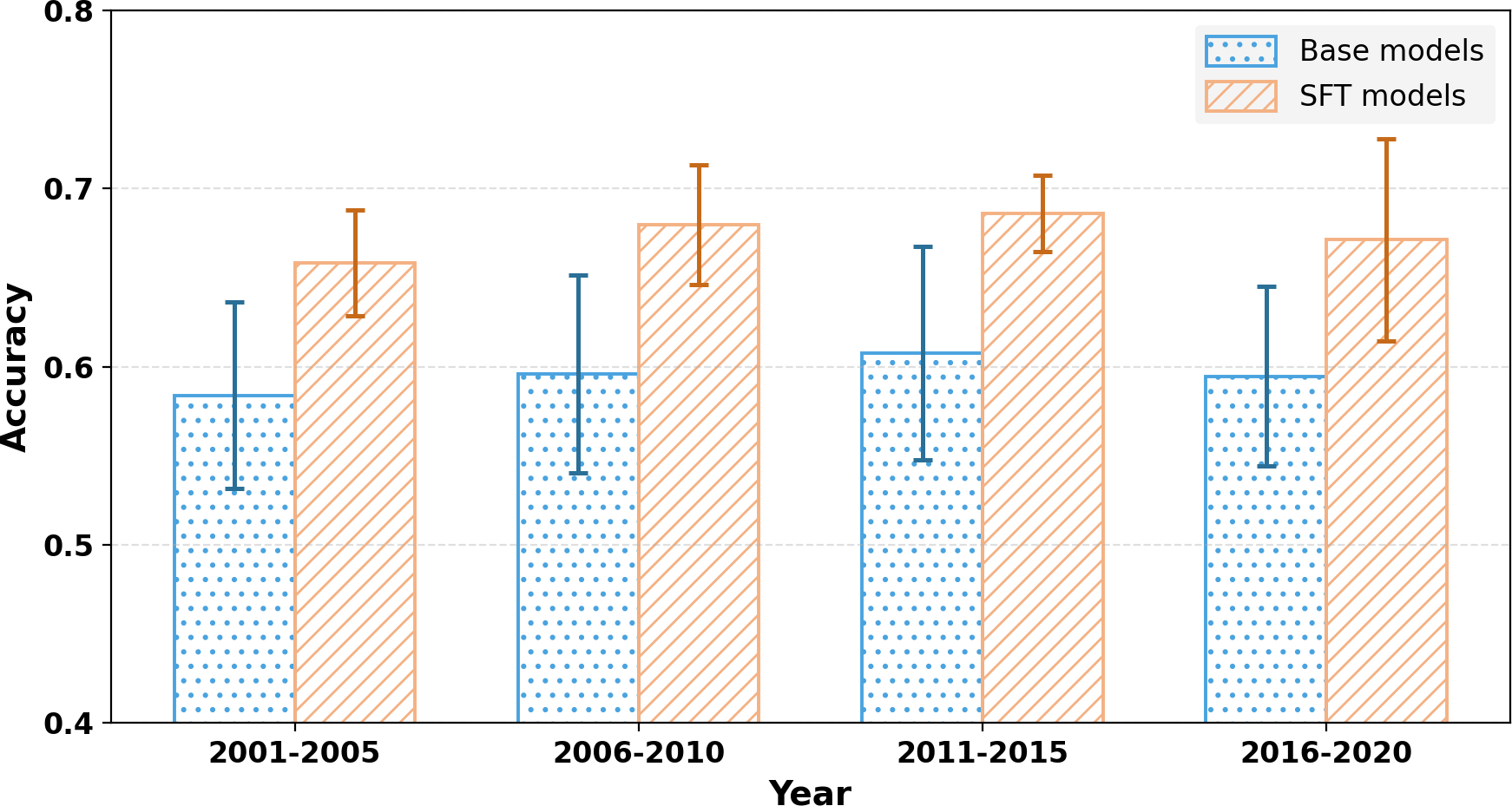}
  \caption{Citation accuracy by publication year. The bar chart compares the average accuracy of Base models (blue, dotted hatch) and SFT models (orange, diagonal hatch) across four five-year intervals from 2001 to 2020. The error bars represent the standard deviation of accuracy across models within each time bin.}
  \vspace{-0.5em}
  \label{fig:citation_accuracy_bar}
\end{figure}

\smallskip
\noindent \textbf{Effect of Publication Period on Prediction Difficulty.}
We further examine whether artifact impact prediction becomes easier or harder for papers published in different time periods. Figure~\ref{fig:citation_accuracy_bar} reports citation prediction accuracy across four five-year publication intervals from 2001 to 2020. We observe that performance remains largely stable across publication periods for both base models and SFT models, with no clear advantage for older papers over newer ones. This suggests that, under our benchmark design, prediction difficulty is not strongly driven by publication era. A likely reason is that models observe only titles and abstracts, without temporal cues such as publication year, citation histories, or early reception signals. As a result, age-related information is intentionally removed, forcing models to rely on intrinsic textual signals and leading to broadly comparable difficulty across time bins. At the same time, SFT models consistently outperform their base counterparts in every interval, indicating that the benefits of training on \ours transfer robustly across different publication periods rather than concentrating on a specific era. Taken together, these findings suggest that temporal variation in publication year has only a limited effect on citation prediction difficulty in our text-only setting.

\section{Conclusion}

We introduce \ours, a multi-dimensional, multi-field benchmark for scientific impact prediction, spanning the Citation, Award, Patent, Media, Code, Dataset, and Model dimensions, as well as fields including Computer Science, Physics, Chemistry, Medicine, and various other disciplines. Our evaluation of 11 LLMs demonstrates the heterogeneous nature of scientific impact: models perform better on Best Paper Award prediction, where textual cues closely align with evaluative criteria, whereas dimensions such as Patent and Media remain more challenging due to latent external factors (e.g., market timing and societal relevance). Task-specific training on \ours proves highly effective, with relatively small models like \modelname{SFT-Qwen3-4B} consistently outperforming larger open-source baselines and even stronger closed-source models. Observed performance patterns across dimensions and fields reveal where textual signals are sufficient and where additional context may be necessary. Overall, \ours provides a rigorous platform for evaluating and improving multi-dimensional, multi-field scientific impact prediction, supporting the development of more effective and generalizable models.

\end{spacing}

\section*{Limitations}
\noindent \textbf{Text Truncation and Scope.} We limit textual input to at most 1,000 words across all dimensions to maintain consistent prompt lengths across instances. Consequently, \ours provides models with a truncated view of papers and long-form artifacts such as extended READMEs or dataset/model cards. Future work could explore long-context models or hierarchical and retrieval-based strategies to more fully exploit artifact content while maintaining scalability.


\smallskip 
\noindent \textbf{Recognition vs. Forecasting.}
\ours is not purely an ex ante forecasting benchmark: some instances may be easier because award-winning, highly cited, or canonical artifacts are already salient in pre-training corpora or public discourse. Temporally controlled test sets or time-bounded pre-training would better isolate true prospective forecasting ability.

\smallskip 
\noindent \textbf{Pairwise Simplification.}
We formulate impact prediction as pairwise binary classification, which normalizes scale differences across dimensions and provides a clean test of discriminative ability. However, it does not capture absolute forecasting or ranking over large candidate pools, which remain important directions for future work.

\section*{Ethical Considerations}
A central ethical concern of this work is Goodhart's Law~\cite{strathern1997improving}. If predictive models of scientific impact are used in high-stakes settings such as funding, hiring, or promotion, researchers may optimize writing, topic choice, or dissemination strategies to satisfy model signals rather than improve intrinsic quality, potentially distorting research behavior and narrowing scientific diversity. We therefore emphasize that \ours and models trained on it are intended strictly as decision-support and filtering tools for human discovery and exploration, not as autonomous systems for scientific evaluation.

Moreover, fine-tuned models may inherit biases in historical data and public signals. For example, awards, media attention, or artifact adoption may systematically favor certain fields or institutions, potentially reinforcing existing inequities in science. While \ours broadens impact beyond citations and spans diverse fields, it does not fully eliminate such structural biases. Users should therefore exercise caution, analyze biases across dimensions and fields, and avoid interpreting predictions as normative judgments of scientific merit.

Overall, we view scientific impact prediction as an inherently subjective and multi-faceted task. Our benchmark is designed to facilitate research into this complexity, not to replace expert judgment. Responsible use of \ours requires maintaining human oversight, transparency about limitations, and restraint in applying model predictions to consequential decisions.


\bibliography{custom}

@article{li2019dataset,
  title={A dataset of publication records for Nobel laureates},
  author={Li, Jichao and Yin, Yian and Fortunato, Santo and Wang, Dashun},
  journal={Scientific Data},
  volume={6},
  number={1},
  pages={33},
  year={2019}
}

@inproceedings{li2019neural,
  title={A neural citation count prediction model based on peer review text},
  author={Li, Siqing and Zhao, Wayne Xin and Yin, Eddy Jing and Wen, Ji-Rong},
  booktitle={EMNLP'19},
  pages={4914--4924},
  year={2019}
}

@inproceedings{hirako2023realistic,
  title={Realistic citation count prediction task for newly published papers},
  author={Hirako, Jun and Sasano, Ryohei and Takeda, Koichi},
  booktitle={Findings of EACL'23},
  pages={1131--1141},
  year={2023}
}

@article{lin2023sciscinet,
  title={SciSciNet: A large-scale open data lake for the science of science research},
  author={Lin, Zihang and Yin, Yian and Liu, Lu and Wang, Dashun},
  journal={Scientific Data},
  volume={10},
  number={1},
  pages={315},
  year={2023}
}

@inproceedings{yang2024navigating,
  title={Navigating Dataset Documentations in AI: A Large-Scale Analysis of Dataset Cards on HuggingFace},
  author={Yang, Xinyu and Liang, Weixin and Zou, James},
  booktitle={ICLR'24},
  year={2024}
}

@article{liang2024systematic,
  title={Systematic analysis of 32,111 AI model cards characterizes documentation practice in AI},
  author={Liang, Weixin and Rajani, Nazneen and Yang, Xinyu and Ozoani, Ezinwanne and Wu, Eric and Chen, Yiqun and Smith, Daniel Scott and Zou, James},
  journal={Nature Machine Intelligence},
  volume={6},
  number={7},
  pages={744--753},
  year={2024}
}

@inproceedings{zhang2025internal,
  title={Internal and External Impacts of Natural Language Processing Papers},
  author={Zhang, Yu},
  booktitle={ACL'25},
  pages={488--494},
  year={2025}
}

@inproceedings{dong2017century,
  title={A century of science: Globalization of scientific collaborations, citations, and innovations},
  author={Dong, Yuxiao and Ma, Hao and Shen, Zhihong and Wang, Kuansan},
  booktitle={KDD'17},
  pages={1437--1446},
  year={2017}
}

@article{xia2023review,
  title={A review of scientific impact prediction: tasks, features and methods},
  author={Xia, Wanjun and Li, Tianrui and Li, Chongshou},
  journal={Scientometrics},
  volume={128},
  number={1},
  pages={543--585},
  year={2023}
}

@inproceedings{dong2015will,
  title={Will this paper increase your h-index? Scientific impact prediction},
  author={Dong, Yuxiao and Johnson, Reid A and Chawla, Nitesh V},
  booktitle={WSDM'15},
  pages={149--158},
  year={2015}
}

@article{yin2022public,
  title={Public use and public funding of science},
  author={Yin, Yian and Dong, Yuxiao and Wang, Kuansan and Wang, Dashun and Jones, Benjamin F},
  journal={Nature Human Behaviour},
  volume={6},
  number={10},
  pages={1344--1350},
  year={2022}
}

@article{radicchi2017quantifying,
  title={Quantifying perceived impact of scientific publications},
  author={Radicchi, Filippo and Weissman, Alexander and Bollen, Johan},
  journal={Journal of Informetrics},
  volume={11},
  number={3},
  pages={704--712},
  year={2017}
}

@article{jin2021scientific,
  title={Scientific prizes and the extraordinary growth of scientific topics},
  author={Jin, Ching and Ma, Yifang and Uzzi, Brian},
  journal={Nature Communications},
  volume={12},
  number={1},
  pages={5619},
  year={2021}
}

@inproceedings{shen2018web,
  title={A Web-scale system for scientific knowledge exploration},
  author={Shen, Zhihong and Ma, Hao and Wang, Kuansan},
  booktitle={ACL'18},
  pages={87--92},
  year={2018}
}

@inproceedings{zhang2023effect,
  title={The effect of metadata on scientific literature tagging: A cross-field cross-model study},
  author={Zhang, Yu and Jin, Bowen and Zhu, Qi and Meng, Yu and Han, Jiawei},
  booktitle={WWW'23},
  pages={1626--1637},
  year={2023}
}

@article{yang2025qwen3,
  title={Qwen3 technical report},
  author={Yang, An and Li, Anfeng and Yang, Baosong and Zhang, Beichen and Hui, Binyuan and Zheng, Bo and Yu, Bowen and Gao, Chang and Huang, Chengen and Lv, Chenxu},
  journal={arXiv preprint arXiv:2505.09388},
  year={2025}
}

@article{grattafiori2024llama,
  title={The llama 3 herd of models},
  author={Grattafiori, Aaron and Dubey, Abhimanyu and Jauhri, Abhinav and Pandey, Abhinav and Kadian, Abhishek and Al-Dahle, Ahmad and Letman, Aiesha and Mathur, Akhil and Schelten, Alan and Vaughan, Alex},
  journal={arXiv preprint arXiv:2407.21783},
  year={2024}
}

@article{priem2022openalex,
  title={OpenAlex: A fully-open index of scholarly works, authors, venues, institutions, and concepts},
  author={Priem, Jason and Piwowar, Heather and Orr, Richard},
  journal={arXiv preprint arXiv:2205.01833},
  year={2022}
}

@misc{huang2023best,
  author={Jeff Huang},
  title={Best Paper Awards in Computer Science (since 1996)},
  year={2023},
  url={https://jeffhuang.com/best_paper_awards}
}

@misc{paperswithcode,
  author={{Papers with Code}},
  title={Links between papers and code},
  year={2019},
  url={https://huggingface.co/datasets/pwc-archive/links-between-paper-and-code}
}

@inproceedings{castillo2007estimating,
  title={Estimating number of citations using author reputation},
  author={Castillo, Carlos and Donato, Debora and Gionis, Aristides},
  booktitle={International Symposium on String Processing and Information Retrieval},
  pages={107--117},
  year={2007}
}

@inproceedings{fu2008models,
  title={Models for predicting and explaining citation count of biomedical articles},
  author={Fu, Lawrence D and Aliferis, Constantin},
  booktitle={AMIA'08},
  pages={222},
  year={2008}
}

@article{ibanez2009predicting,
  title={Predicting citation count of Bioinformatics papers within four years of publication},
  author={Ib{\'a}{\~n}ez, Alfonso and Larra{\~n}aga, Pedro and Bielza, Concha},
  journal={Bioinformatics},
  volume={25},
  number={24},
  pages={3303--3309},
  year={2009}
}

@inproceedings{chakraborty2014towards,
  title={Towards a stratified learning approach to predict future citation counts},
  author={Chakraborty, Tanmoy and Kumar, Suhansanu and Goyal, Pawan and Ganguly, Niloy and Mukherjee, Animesh},
  booktitle={JCDL'14},
  pages={351--360},
  year={2014}
}

@inproceedings{xiao2016modeling,
  title={On Modeling and Predicting Individual Paper Citation Count over Time.},
  author={Xiao, Shuai and Yan, Junchi and Li, Changsheng and Jin, Bo and Wang, Xiangfeng and Yang, Xiaokang and Chu, Stephen M and Zha, Hongyuan},
  booktitle={IJCAI'16},
  pages={2676--2682},
  year={2016}
}

@article{yuan2018modeling,
  title={Modeling and predicting citation count via recurrent neural network with long short-term memory},
  author={Yuan, Sha and Tang, Jie and Zhang, Yu and Wang, Yifan and Xiao, Tong},
  journal={arXiv preprint arXiv:1811.02129},
  year={2018}
}

@inproceedings{ren2020starin,
  title={Starin: An approach to predict the popularity of github repository},
  author={Ren, Leiming and Shan, Shimin and Xu, Xiujuan and Liu, Yu},
  booktitle={International Conference of Pioneering Computer Scientists, Engineers and Educators},
  pages={258--273},
  year={2020}
}

@inproceedings{koch2021reduced,
  title={Reduced, Reused and Recycled: The Life of a Dataset in Machine Learning Research},
  author={Koch, Bernard and Denton, Emily and Hanna, Alex and Foster, Jacob Gates},
  booktitle={NeurIPS'21},
  year={2021}
}

@inproceedings{koch2024fault,
  title={The Fault in Our Stars: An Analysis of GitHub Stars as an Importance Metric for Web Source Code},
  author={Koch, Simon and Klein, David and Johns, Martin},
  booktitle={Workshop on Measurements, Attacks, and Defenses for the Web},
  year={2024}
}

@inproceedings{zhao2025words,
  title={From Words to Worth: Newborn Article Impact Prediction with LLM},
  author={Zhao, Penghai and Xing, Qinghua and Dou, Kairan and Tian, Jinyu and Tai, Ying and Yang, Jian and Cheng, Ming-Ming and Li, Xiang},
  booktitle={AAAI'25},
  pages={1183--1191},
  year={2025}
}

@article{lu2025newborn,
  title={From Newborn to Impact: Bias-Aware Citation Prediction},
  author={Lu, Mingfei and Wu, Mengjia and Xu, Jiawei and Li, Weikai and Liu, Feng and Ding, Ying and Sun, Yizhou and Lu, Jie and Zhang, Yi},
  journal={arXiv preprint arXiv:2510.19246},
  year={2025}
}

@inproceedings{gu2024impact4cast,
  title={Impact4cast: forecasting high-impact research topics via machine learning on evolving knowledge graphs},
  author={Gu, Xuemei and Krenn, Mario},
  booktitle={ICML 2024 AI for Science Workshop},
  year={2024}
}

@article{rezaee2025fos,
  title={FOS: A Large-Scale Temporal Graph Benchmark for Scientific Interdisciplinary Link Prediction},
  author={Rezaee, Kiyan and Ziabakhsh, Morteza and Nikfarjam, Niloofar and Ghassemi, Mohammad M and Jouryabi, Yazdan Rezaee and Eskandari, Sadegh and Lashgari, Reza},
  journal={arXiv preprint arXiv:2511.18631},
  year={2025}
}

@article{yang2025qwen2,
  title={Qwen2.5 Technical Report},
  author={An Yang and Baosong Yang and Beichen Zhang and Binyuan Hui and Bo Zheng and Bowen Yu and Chengyuan Li and Dayiheng Liu and Fei Huang and Haoran Wei},
  journal={arXiv preprint arXiv:2412.15115},
  year={2024}
}

@article{blakeman2025nemotron,
  title={Nemotron 3 Nano: Open, Efficient Mixture-of-Experts Hybrid Mamba-Transformer Model for Agentic Reasoning},
  author={Blakeman, Aaron and Grattafiori, Aaron and Basant, Aarti and Gupta, Abhibha and Khattar, Abhinav and Renduchintala, Adi and Vavre, Aditya and Shukla, Akanksha and Bercovich, Akhiad and Ficek, Aleksander},
  journal={arXiv preprint arXiv:2512.20848},
  year={2025}
}

@misc{mistral2025ministral3,
  title={Ministral 3: Strong edge-ready AI},
  author={{Mistral AI Team}},
  year={2025},
  url={https://mistral.ai/news/mistral-3}
}

@article{achiam2023gpt,
  title={Gpt-4 technical report},
  author={Achiam, Josh and Adler, Steven and Agarwal, Sandhini and Ahmad, Lama and Akkaya, Ilge and Aleman, Florencia Leoni and Almeida, Diogo and Altenschmidt, Janko and Altman, Sam and Anadkat, Shyamal},
  journal={arXiv preprint arXiv:2303.08774},
  year={2023}
}

@misc{openai2025o3o4mini,
  title={OpenAI o3 and o4-mini System Card},
  author={OpenAI},
  year={2025},
  url={https://cdn.openai.com/pdf/2221c875-02dc-4789-800b-e7758f3722c1/o3-and-o4-mini-system-card.pdf}
}

@misc{anthropic2025claudehaiku45,
  title={System Card: Claude Haiku 4.5},
  author={Anthropic},
  year={2025},
  url={https://assets.anthropic.com/m/99128ddd009bdcb/Claude-Haiku-4-5-System-Card.pdf}
}

@article{sinatra2016quantifying,
  title={Quantifying the evolution of individual scientific impact},
  author={Sinatra, Roberta and Wang, Dashun and Deville, Pierre and Song, Chaoming and Barab{\'a}si, Albert-L{\'a}szl{\'o}},
  journal={Science},
  volume={354},
  number={6312},
  pages={aaf5239},
  year={2016}
}

@article{wang2013quantifying,
  title={Quantifying long-term scientific impact},
  author={Wang, Dashun and Song, Chaoming and Barab{\'a}si, Albert-L{\'a}szl{\'o}},
  journal={Science},
  volume={342},
  number={6154},
  pages={127--132},
  year={2013}
}

@misc{mdpi,
  author={MDPI},
  title={MDPI Awards},
  year={2025},
  url={https://www.mdpi.com/awards}
}

@article{mitsis2022nobel,
  title={The Nobel Prize time gap},
  author={Mitsis, Pandelis},
  journal={Humanities and Social Sciences Communications},
  volume={9},
  number={1},
  pages={407},
  year={2022}
}

@article{szomszor2022overton,
  title={Overton: A bibliometric database of policy document citations},
  author={Szomszor, Martin and Adie, Euan},
  journal={Quantitative Science Studies},
  volume={3},
  number={3},
  pages={624--650},
  year={2022}
}

@misc{github,
  author={GitHub},
  title={GitHub REST API documentation},
  year={2022},
  url={https://docs.github.com/en/rest?apiVersion=2022-11-28}
}

@inproceedings{ammar2018construction,
  title={Construction of the literature graph in semantic scholar},
  author={Ammar, Waleed and Groeneveld, Dirk and Bhagavatula, Chandra and Beltagy, Iz and Crawford, Miles and Downey, Doug and Dunkelberger, Jason and Elgohary, Ahmed and Feldman, Sergey and Ha, Vu and others},
  booktitle={NAACL'18},
  pages={84--91},
  year={2018}
}

@article{strathern1997improving,
  title={‘Improving ratings’: audit in the British University system},
  author={Strathern, Marilyn},
  journal={European Review},
  volume={5},
  number={3},
  pages={305--321},
  year={1997}
}

@inproceedings{beltagy2019scibert,
  title={SciBERT: A pretrained language model for scientific text},
  author={Beltagy, Iz and Lo, Kyle and Cohan, Arman},
  booktitle={EMNLP'19},
  pages={3615--3620},
  year={2019}
}

@inproceedings{zhang2024comprehensive,
  title={A comprehensive survey of scientific large language models and their applications in scientific discovery},
  author={Zhang, Yu and Chen, Xiusi and Jin, Bowen and Wang, Sheng and Ji, Shuiwang and Wang, Wei and Han, Jiawei},
  booktitle={EMNLP'24},
  pages={8783--8817},
  year={2024}
}

@inproceedings{sayyadi2009futurerank,
  title={Futurerank: Ranking scientific articles by predicting their future pagerank},
  author={Sayyadi, Hassan and Getoor, Lise},
  booktitle={SDM'09},
  pages={533--544},
  year={2009}
}

@article{gao2025science,
  title={Science hierarchography: Hierarchical organization of science literature},
  author={Gao, Muhan and Shah, Jash and Wang, Weiqi and Huang, Kuan-Hao and Khashabi, Daniel},
  journal={arXiv preprint arXiv:2504.13834},
  year={2025}
}

\appendix
\section{Comparison across Fields}
\label{app:radar_field}
\begin{figure}[h]
  \centering
  \includegraphics[width=\columnwidth]{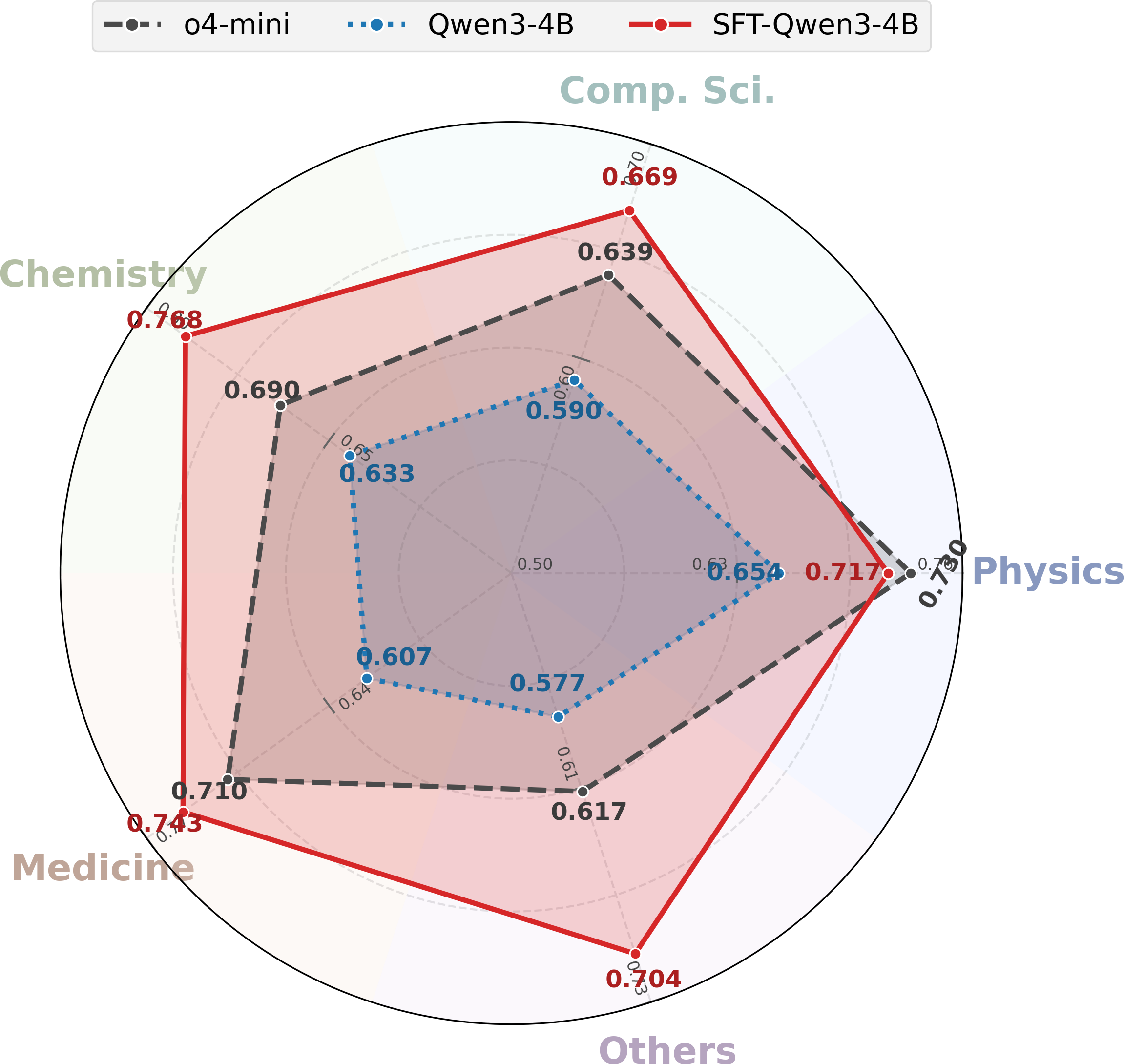}
  \caption{Performance of \modelname{o4-mini}, off-the-shelf \modelname{Qwen3-4B}, and supervised fine-tuned \modelname{Qwen3-4B} across scientific fields on \ours.
  SFT substantially enhances a 4B open-weight model's ability to predict scientific impact across all fields, enabling it to rival or surpass stronger closed-source models.}
  \label{fig:radar_field}
  \vspace{-0.5em}
\end{figure}

Figure~\ref{fig:radar_field} provides a field-wise view of the same three representative models examined in Figure~\ref{fig:radar_metric}, illustrating how performance varies across scientific fields and how SFT affects cross-field generalization.

\section{Prompts}
\label{app:prompt}

\subsection{Citation}
\colorbox{mygray}{%
\parbox{0.97\columnwidth}{%
\small
\textbf{System:} You are an impartial judge deciding which of two research papers has more citations. Your reply must be exactly one sentence and must be one of these two options: \\
\hspace*{1em}-- Paper A has more citations \\
\hspace*{1em}-- Paper B has more citations \\
You are not allowed to output anything else---no explanations, no extra words. \\[0.5em]
\textbf{User:} Paper A: \texttt{<artifact text for A>} \\[0.25em]
Paper B: \texttt{<artifact text for B>} \\[0.5em]
Based solely on the information above, which paper do you think has more citations? \\
Reply with exactly one sentence following the system instruction.
}%
}

\subsection{Best Paper Award (CS Conferences)}
\colorbox{mygray}{%
\parbox{0.97\columnwidth}{%
\small
\textbf{System:} You are an impartial paper reviewer. Given the titles and abstracts of two papers, identify which paper won the Best Paper award. Your reply must be exactly one sentence and must be one of these two options: \\
\hspace*{1em}-- Paper A won the best paper award. \\
\hspace*{1em}-- Paper B won the best paper award. \\
You are not allowed to output anything else---no explanations, no extra words. \\[0.5em]
\textbf{User:} Paper A: \texttt{<artifact text for A>} \\[0.25em]
Paper B: \texttt{<artifact text for B>} \\[0.5em]
Based on the information above, which paper should win the Best Paper award? \\
Reply with exactly one sentence following the system instruction.
}%
}

\subsection{Best Paper Award (MDPI Journals)}
\colorbox{mygray}{%
\parbox{0.97\columnwidth}{%
\small
\textbf{System:} You are an impartial judge deciding which of two MDPI papers won the MDPI Best Paper Award. Your reply must be exactly one sentence and must be one of these two options: \\
\hspace*{1em}-- Paper A won the MDPI Best Paper Award \\
\hspace*{1em}-- Paper B won the MDPI Best Paper Award \\
You are not allowed to output anything else---no explanations, no extra words. \\[0.5em]
\textbf{User:} Paper A: \texttt{<artifact text for A>} \\[0.25em]
Paper B: \texttt{<artifact text for B>} \\[0.5em]
Based on the information above, which paper won the MDPI Best Paper Award? \\
Reply with exactly one sentence following the system instruction.
}%
}

\subsection{Nobel Prize}
\colorbox{mygray}{%
\parbox{0.97\columnwidth}{%
\small
\textbf{System:} You are an impartial judge deciding which of two research papers is the Nobel prize-winning paper. Your reply must be exactly one sentence and must be one of these two options: \\
\hspace*{1em}-- Paper A is the Nobel prize-winning paper. \\
\hspace*{1em}-- Paper B is the Nobel prize-winning paper. \\
You are not allowed to output anything else---no explanations, no extra words. \\[0.5em]
\textbf{User:} Paper A: \texttt{<artifact text for A>} \\[0.25em]
Paper B: \texttt{<artifact text for B>} \\[0.5em]
Based on the information above, which paper is the Nobel prize-winning paper? \\
Reply with exactly one sentence following the system instruction.
}%
}

\subsection{Patent}
\colorbox{mygray}{%
\parbox{0.97\columnwidth}{%
\small
\textbf{System:} You are an impartial judge deciding which of two research papers would be cited in more patents. Your reply must be exactly one sentence and must be one of these two options: \\
\hspace*{1em}-- Paper A could be cited in more patents. \\
\hspace*{1em}-- Paper B could be cited in more patents. \\
You are not allowed to output anything else---no explanations, no extra words. \\[0.5em]
\textbf{User:} Paper A: \texttt{<artifact text for A>} \\[0.25em]
Paper B: \texttt{<artifact text for B>} \\[0.5em]
Based on the information above, which paper could be cited in more patents? \\
Reply with exactly one sentence following the system instruction.
}%
}

\subsection{Media}
\colorbox{mygray}{%
\parbox{0.97\columnwidth}{%
\small
\textbf{System:} You are an impartial judge deciding which of two research papers would be cited in more media mentions. Your reply must be exactly one sentence and must be one of these two options: \\
\hspace*{1em}-- Paper A could get more media mentions. \\
\hspace*{1em}-- Paper B could get more media mentions. \\
You are not allowed to output anything else---no explanations, no extra words. \\[0.5em]
\textbf{User:} Paper A: \texttt{<artifact text for A>} \\[0.25em]
Paper B: \texttt{<artifact text for B>} \\[0.5em]
Based on the information above, which paper could get more media mentions? \\
Reply with exactly one sentence following the system instruction.
}%
}

\subsection{Code}
\colorbox{mygray}{%
\parbox{0.97\columnwidth}{%
\small
\textbf{System:} You are an impartial judge deciding which of two GitHub repositories has more stars. Your reply must be exactly one sentence and must be one of these two options: \\
\hspace*{1em}-- GitHub repo A has more stars. \\
\hspace*{1em}-- GitHub repo B has more stars. \\
You are not allowed to output anything else---no explanations, no extra words. \\[0.5em]
\textbf{User:} GitHub repo A README: \texttt{<artifact text for A>} \\[0.25em]
GitHub repo B README: \texttt{<artifact text for B>} \\[0.5em]
Based on the information above, which repository has more stars? \\
Reply with exactly one sentence following the system instruction.
}%
}

\subsection{Dataset}
\colorbox{mygray}{%
\parbox{0.97\columnwidth}{%
\small
\textbf{System:} You are an impartial judge deciding which of two Hugging Face datasets has more downloads. Your reply must be exactly one sentence and must be one of these two options: \\
\hspace*{1em}-- Dataset A has more downloads. \\
\hspace*{1em}-- Dataset B has more downloads. \\
You are not allowed to output anything else---no explanations, no extra words. \\[0.5em]
\textbf{User:} Dataset A: \texttt{<artifact text for A>} \\[0.25em]
Dataset B: \texttt{<artifact text for B>} \\[0.5em]
Based on the information above, which dataset has more downloads? \\
Reply with exactly one sentence following the system instruction.
}%
}

\subsection{Model}
\colorbox{mygray}{%
\parbox{0.97\columnwidth}{%
\small
\textbf{System:} You are an impartial judge deciding which of two Hugging Face models has more downloads. Your reply MUST be exactly one sentence and must be one of these two options: \\
\hspace*{1em}-- Model A has more downloads. \\
\hspace*{1em}-- Model B has more downloads. \\
You are not allowed to output anything else---no explanations, no extra words. \\[0.5em]
\textbf{User:} Model A: \texttt{<artifact text for A>} \\[0.25em]
Model B: \texttt{<artifact text for B>} \\[0.5em]
Based on the information above, which model has more downloads? \\
Reply with exactly one sentence following the system instruction.
}%
}

\begin{table*}[!htbp]
\centering
\small
\renewcommand{\arraystretch}{1.2}
\setlength{\tabcolsep}{6pt}
\scalebox{0.88}{
\begin{tabular}{lcccccccl}
\toprule
 & \textbf{Citation} & \textbf{Award} & \textbf{Patent} & \textbf{Media} & \textbf{Code} & \textbf{Dataset} & \textbf{Model} & \cellcolor{black!10} \textbf{Average} \\
\midrule
\modelname{SFT-SciBERT} & 0.568 & 0.637 & 0.541 & 0.562 & 0.504 & 0.558 & 0.531 & \cellcolor{black!10} 0.557 \\
\midrule
\modelname{LLaMA-3.2-3B} & 0.534 & 0.539 & 0.534 & 0.517 & 0.513 & 0.526 & 0.548 & \cellcolor{black!10} 0.530 \\
\modelname{Qwen3-4B} & 0.630 & 0.680 & 0.549 & 0.587 & 0.573 & 0.560 & 0.541 & \cellcolor{black!10} 0.589 \\
\midrule
\modelname{SFT-LLaMA-3.2-3B} & 0.653 & 0.806 & 0.629 & 0.697 & 0.618 & 0.618 & 0.625 & \cellcolor{black!10} 0.664 \\
\modelname{SFT-Qwen3-4B} & 0.699 & 0.837 & 0.640 & 0.720 & 0.626 & 0.630 & 0.644 & \cellcolor{black!10} 0.685 \\
\bottomrule
\end{tabular}
}
\caption{Comparison between an encoder baseline (\modelname{SFT-SciBERT}) and LLM methods across impact dimensions.}
\label{tab:encoder_baselines}
\end{table*}

\section{Implementation Details for Supervised Fine-tuning}
\label{app:sft-details}

Both \modelname{Qwen3-4B} and \modelname{LLaMA-3.2-3B} are fine-tuned under an identical training configuration, differing only in the base model checkpoint and the prompt template.
Each training instance consists of a single instruction-following prompt formatted as described in Appendix~\ref{app:prompt}, with a binary forced-choice output. We use the following hyperparameters:

\smallskip
\smallskip
\noindent
\colorbox{mygray}{%
\parbox{0.97\columnwidth}{%
\small
\hspace*{1em}-- Learning rate: \texttt{2e-5} \\
\hspace*{1em}-- Epochs: \texttt{1} (\modelname{Qwen3-4B}), \texttt{3} (\modelname{LLaMA-3.2-3B}) \\
\hspace*{1em}-- Per-device batch size: \texttt{8} (train / eval) \\
\hspace*{1em}-- Gradient accumulation steps: \texttt{2} \\
\hspace*{1em}-- Effective batch size: \texttt{64} \\
\hspace*{1em}-- Learning rate schedule: cosine \\
\hspace*{1em}-- Warmup ratio: \texttt{0.1}
}%
}

\smallskip
\smallskip
We set the maximum input length to 4,096 tokens, truncating longer inputs. Mixed-precision training with bf16 is enabled, and FlashAttention with SDPA is used to improve memory efficiency. We adopt DeepSpeed ZeRO Stage 2 for memory optimization and enable expandable CUDA memory segments to mitigate memory fragmentation during long-context training. Evaluation is performed on the validation split every 500 training steps.

\begin{table*}[t]
\centering
\small
\setlength{\tabcolsep}{6pt}
\renewcommand{\arraystretch}{1.15}
\scalebox{0.88}{
\begin{tabular}{l|cc|cc|cc}
\toprule
& \multicolumn{2}{c|}{\textbf{Code}} & \multicolumn{2}{c|}{\textbf{Dataset}} & \multicolumn{2}{c}{\textbf{Model}} \\
& $\rm Acc_{After}$ & $\Delta\rm Acc$ & $\rm Acc_{After}$ & $\Delta\rm Acc$ & $\rm Acc_{After}$ & $\Delta\rm Acc$ \\
\midrule
\modelname{LLaMA-3.2-3B} & 0.511 & $-$0.002 & 0.518 & $-$0.008 & 0.541 & $-$0.007 \\
\modelname{LLaMA-3-8B} & 0.545 & $-$0.002 & 0.543 & $-$0.006 & 0.524 & $-$0.010 \\
\modelname{LLaMA-3.1-8B} & 0.511 & $-$0.014 & 0.525 & $-$0.009 & 0.533 & $-$0.002 \\
\modelname{Qwen3-4B} & 0.566 & $-$0.007 & 0.591 & $-$0.009 & 0.550 & $+$0.009 \\
\modelname{Qwen2.5-7B} & 0.554 & $-$0.009 & 0.588 & $-$0.002 & 0.560 & 0.000 \\
\modelname{Qwen2.5-14B} & 0.565 & $-$0.012 & 0.548 & $-$0.013 & 0.562 & 0.000 \\
\modelname{Ministral-3-3B} & 0.517 & $-$0.025 & 0.502 & $-$0.001 & 0.519 & 0.000 \\
\modelname{Nemotron-3-Nano-30B} & 0.533 & $+$0.005 & 0.559 & $-$0.002 & 0.541 & $-$0.008 \\
\modelname{SFT-LLaMA-3.2-3B} & 0.619 & $+$0.001 & 0.619 & $+$0.001 & 0.628 & $+$0.003 \\
\modelname{SFT-Qwen3-4B} & 0.636 & $+$0.010 & 0.628 & $-$0.002 & 0.644 & 0.000 \\
\bottomrule
\end{tabular}
}
\caption{Leakage audit after removing explicit popularity cues from the text used in the Code, Dataset, and Model dimensions. $\rm Acc_{After}$: Accuracy after removing explicit popularity cues. $\Delta\rm Acc = Acc_{After} - Acc_{Before}$: Performance change relative to the original input text.}
\vspace{-0.5em}
\label{tab:leakage_audit}
\end{table*}

\section{Encoder Baseline}
\label{app:encoder-baselines}

To contextualize the difficulty of \ours, we additionally train an encoder-based classifier, \modelname{SciBERT} \cite{beltagy2019scibert}, on the same aggregated training split used for SFT. We follow a standard binary classification setup: the input is the concatenation of the task definition and the paired-artifact text, and a linear classification head is applied to the \texttt{[CLS]} representation to predict whether artifact A or artifact B has higher impact. Because BERT-style encoders are limited to 512 tokens, longer examples are truncated by allocating the token budget evenly across the two artifacts.

As shown in Table~\ref{tab:encoder_baselines}, \modelname{SFT-SciBERT} reaches an average accuracy of 0.557 and outperforms untuned \modelname{LLaMA-3.2-3B} on five of the seven impact dimensions. This indicates that \ours provides a meaningful supervision signal even for relatively small models.
At the same time, \modelname{Qwen3-4B} still surpasses \modelname{SFT-SciBERT} on every dimension, and the gap widens further for \modelname{SFT-Qwen3-4B}. To summarize, simple supervised encoders are already non-trivial contenders, but strong LLMs still deliver substantial additional gains.

\section{Leakage Audit for Code, Dataset, and Model}
\label{app:leakage-audit}

For the Code, Dataset, and Model dimensions, the textual input consists only of the main descriptive text on the artifact page, namely the repository README, Hugging Face dataset card, or Hugging Face model card. Structured popularity counters such as GitHub stars / forks and Hugging Face downloads are used only to define the ground-truth labels, not as input features.

To further test whether explicit popularity cues leak into the text, we implement a de-leakage pipeline that removes: (1) markdown / HTML badges and images, including badge links; (2) direct popularity counts and common numeric patterns such as ``\textit{1.2k stars}'', ``\textit{downloads: 320K}'', or ``\textit{forks 56}''; (3) popularity descriptors and usage-count claims such as ``\textit{trending}'', ``\textit{popular}'', or ``\textit{used by X projects / companies / papers}''; (4) Hugging Face statistics templates involving views, likes, or downloads; and (5) lines dominated by badge or stat tokens.
Table~\ref{tab:leakage_audit} reports model performance after cleaning, along with the corresponding performance changes. Results remain highly stable across models, and the overall ranking of model families is unchanged. This suggests that performance on these three dimensions is not primarily driven by trivial leakage from badges or popularity-count strings.

\end{document}